# Multilingual Text Classification for Dravidian Languages

Xiaotian Lin[1], Nankai Lin[1], Kanoksak Wattanachote[1], Shengyi Jiang[1,2] (✉), Lianxi Wang[1,2] (✉)

1. School of Information Science and Technology, Guangdong University of Foreign Studies Guangzhou
2. Guangzhou Key Laboratory of Multilingual Intelligent Processing, Guangdong University of Foreign Studies, Guangzhou

jiangshengyi@163.com, wanglianxi@gdufs.edu.cn

**Abstract.** As the fourth largest language family in the world, the Dravidian languages have become a research hotspot in natural language processing (NLP). Although the Dravidian languages contain a large number of languages, there are relatively few public available resources. Besides, text classification task, as a basic task of natural language processing, how to combine it to multiple languages in the Dravidian languages, is still a major difficulty in Dravidian Natural Language Processing. Hence, to address these problems, we proposed a multilingual text classification framework for the Dravidian languages. On the one hand, the framework used the LaBSE pre-trained model as the base model. Aiming at the problem of text information bias in multi-task learning, we propose to use the MLM strategy to select language-specific words, and used adversarial training to perturb them. On the other hand, in view of the problem that the model cannot well recognize and utilize the correlation among languages, we further proposed a language-specific representation module to enrich semantic information for the model. The experimental results demonstrated that the framework we proposed has a significant performance in multilingual text classification tasks with each strategy achieving certain improvements.

## 1. Introduction

Dravidian languages are the common terminology used to represent the South Indian languages, which consist of around 26 languages. Out of these 26 Dravidian languages, Tamil, Malayalam, and Kannada are regarded as official languages and have been spoken by around 220 million people in the Indian subcontinent, Singapore, and Sri Lanka. As the fourth largest languages in the world, although scholars have carried out targeted research on it, there are mainly the following problems in it.

(1) Existing researches on Dravidian languages mainly focus on processing text in one certain language. Nevertheless, Dravidian languages include multiple languages, and technology for one certain language cannot be applied to other languages well, even if these languages are in the same language family.

(2) As far as we know, due to the existence of shared tokens among Dravidian languages, there is a certain correlation among Dravidian languages. Nevertheless, the existing models focus on studying monolingual without considering the correlation among the Dravidian languages. These researches

---

cannot effectively promote the multilingual research of the entire Dravidian language.
(3) Most of the existing works utilize language features to improve the effectiveness of the model, such as affixes, syntactic structure, and so on. Because of the differences in grammar among different languages, these methods are difficult to transfer to other languages.
(4) Multi-task learning is an important technology in natural language processing. It can solve the problem of scarcity of annotation resources for each task by combining multiple tasks. There are some studies that apply multi-task learning to multi-language research, so that information between multiple languages can be shared. However, there are soly slightly researchers, whose their researches apply multi-task learning to deal with multilingual tasks in the Dravidian language.

Hence, we aimed to apply multi-task learning technology for the Dravidian language to address these problems. Multi-task Learning (MTL) proposed to learn shared information among multiple related tasks and obtain better performance than learning each task independently. Since multi-task learning utilizes potential correlations among related tasks to extract common features and yield performance gains, it has been widely used in text classification tasks. For example, Zhao et al. (2020) proposed to utilize multi-task learning for text classification, which is composed of a shared encoder, a multi-label classification decoder, and a hierarchical categorization decoder. However, because multi-task learning usually shares the parameters of the general presentation layer among different task, it will cause the problem of text information bias when processing the multilingual classification tasks, resulting in the model only performing well in individual languages. Therefore, how to integrate more task learning for multilingual text classification tasks, is still a major difficulty in NLP.

To address these issues above, based on multi-task learning and adversarial training, we proposed a multilingual text classification framework for the Dravidian languages. On the one hand, the framework uses the LaBSE pre-trained model as the base model. Aiming at the problem of text information bias in multi-task learning, we propose to use the MLM strategy to select language-specific words, and use adversarial training to perturb them. On the other hand, in view of the problem that the model cannot well recognize and utilize the correlation among languages, we further propose a language representation module to enrich semantic information for the model.

The main contributions of this paper are as follows:
(1) We proposed a multilingual text classification framework to better handle multilingual text classification tasks.
(2) We proposed language-specific word extraction technology based on MLM strategy to extract language-specific words.
(3) We proposed to perturb the language information to solve the problem of language information bias in multilingual text classification.
(4) We proposed an innovative method to extract the knowledge about the correlation among languages into the model.
(5) The framework we proposed has a significant performance in multilingual text classification tasks with each strategy achieving certain improvements.

## 2. Related work

## 2.1 Multilingual text classification

Multilingual text classification is one of the research hotspots in the field of natural language processing. Compared with single-language text classification, there are two unresolved difficulties: (1) It is difficult to share and uniformly express the semantic space of different languages. (2) The existing methods are mostly for single language texts while it has low adaptability to Multilingual texts. To address these problems, some researchers put forward some valuable approaches. Liu et al. (2018) proposed to use automatic associative memory with multilingual data fusion to realize multilingual short text classification tasks. Meng et al. (2019) proposed a model based on LDA and Bi-LSTM-CNN to solve the problem of multilingual short text classification, and used topic vectors and word vectors to extract text information in each language. In this vein, their proposed ideas solved the problem of the scarcity in short text features to a certain extent. Meng et al. (2020) proposed a multilingual text classification model combining Bi-LSTM and CNN to extract text features and obtain deeper text representation in various languages. Meng (2018) used multilingual text feature conversion and fusion strategies to solve the domain adaptability of the classifier in different languages, and employed deep learning strategies to improve the accuracy of the classifiers. Groenwold et al. (2020) evaluated the effects of multiple pre-trained language models on multilingual classification tasks. Kazhuparambil and Kaushik (2020) proposed to use the XLM model to automatically classify YouTube comments mixed in English and Malay, which achieves the best results. Mishra et al. (2020) developed to leverage a variety of transformed models for different data sets in the TRAC2020 evaluation task and fine-tuned them. They also proposed joint label classification and multi-language joint training methods to improve classification performance for label marginalization problems.

Although these multilingual text classification methods have attracted increasing research interests in recent years, there are no publicly available corpus and targeted researches for multilingual China-related news classification.

## 2.2 Researches for Dravidian languages

So far, there are relatively few researches on text classification in term of the Dravidian languages and the lack of publicly available corpus. As far as we know, the first evaluation task for hate speech/ offensive content detection and sentiment analysis for the Dravidian languages were all first proposed in FIRE2020.

In offensive Language identification tasks, Sai and Sharma (2020) proposed a novel method using translation and transliteration method, which can obtain better results from fine-tuning and integrated multi-language converter networks such as XLM-RoBERTa and mBERT. Hande et al. (2021) proposed to generate a pseudo-labels dataset called CMTRA to increase the amount of training data for the language models and fine-tune several recent pre-trained language models on this newly constructed dataset. In addition, through using TF-IDF vectors and character-level n-grams as features. Veena et al. (2020) developed and evaluated four systems for processing Malayalam, including logistic regression,

XGBoost, long and short-term memory networks, and attention networks. As for the sentiment analysis tasks, to overcome the code-mixed and grammatical irregularities problems, Chakravarthi et al. (2020) created a gold standard Tamil-English code-switched, sentiment-annotated corpus. Sachin Kumar et al. (2018) focused on providing a comparative study for identifying sentiment of Malayalam tweets using deep learning methods such as convolutional neural net (CNN), long short-term memory units (LSTM). Sun and Zhou (2020) used the hidden state of XLM-Roberta to extract semantic information. They proposed a new model by extracting the output of the top hidden layer in XLM-Roberta and providing them as input to the convolutional neural network, and finally connecting them to obtain better results.

## 2.3 Adversarial training in natural language processing

Adversarial training refers to the method of constructing adversarial samples and mixing them with the original samples to train the model, which can improve the generalization performance of the original examples. Previous work mainly used adversarial training in the image field. For natural language processing, there are several targeted researches.

Gao et al. (2018) proposed to generate small adversarial perturbations for the original samples ina black-box setting, making the model misclassify the text sample. In particular, their method includes two steps: (1) Finding the most important words for modification through a scoring strategy, which causes the deep classifier to make an incorrect prediction. (2) By applying a simple character-level conversion to word ranked highest to make the edit distance of the disturbance minimized. Miyato et al. (2017) first proposed to use adversarial training for word embedding in text input. However, this method is lack of interpretability. In order not to affect the performance of the model, while generating adversarial text for perturbation, Sato et al. (2018) proposed to limit the directions of perturbations toward the existing words in the word embeddings.

## 3. Framework

In this paper, we proposed a multilingual text classification framework based on multi-task learning and adversarial training. As shown in Figure 1, the framework contains four components: (1) Text general representation module; (2) Language-specific words extraction module; (3) Language information perturbation module; (4) Language-specific representation module.

The framework uses the text general representation module to extract the vector representation of the text. At the same time, in the language-specific words extraction module, the framework utilizes MLM strategy to identify important language-specific words in sentences. Next, the language information perturbation module injects different degrees of noise into the embeddings of language-specific words and other words to perturb the generalization of the model. In order to fully learn the general semantic information and the correlation among languages, we further spliced the language descriptor with the sentence vector $S_i$ output by the general representation module.

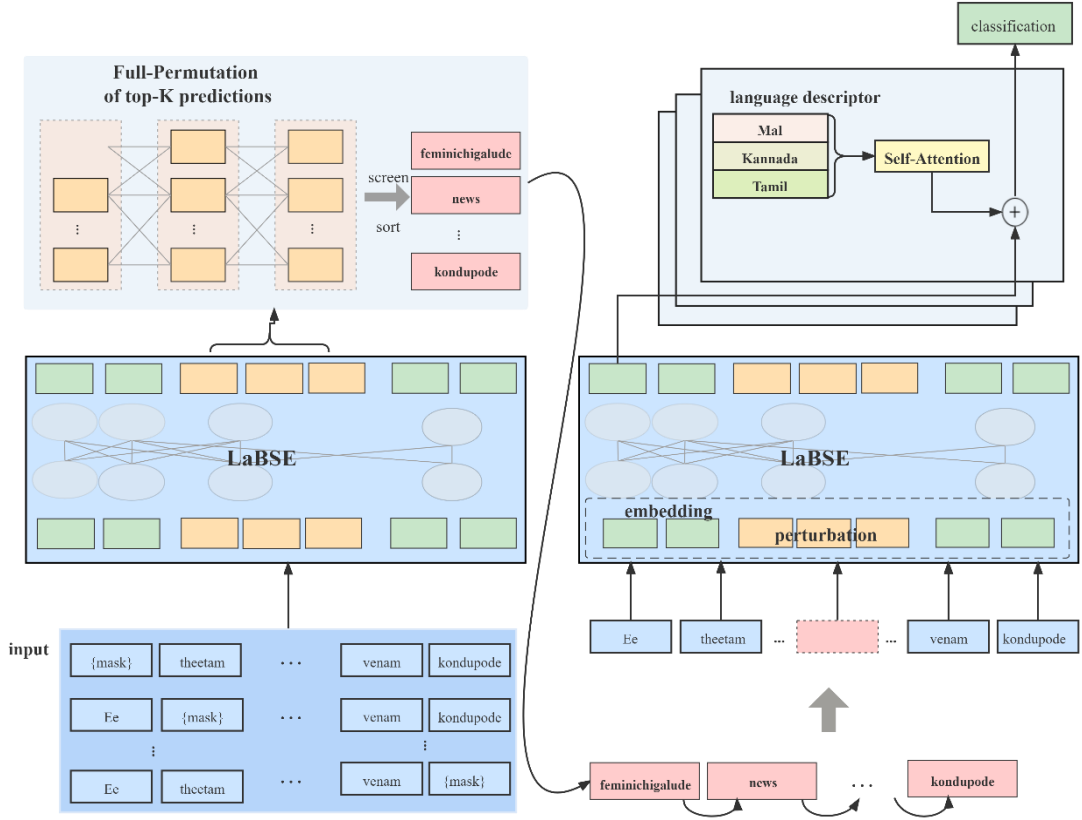

Figure 1. The framework structure.

## 3.1 Text general representation

LaBSE (Language-agnostic BERT Sentence Encoder)[1] presented by Feng et al. (2020) is a BERT-based model trained on 17 billion monolingual sentences and 6 billion bilingual sentence pairs, resulting in a model that is effective even on low-resource languages for which there is no data available during training. What's more, unlike BERT, LaBSE removes the NSP (Next Sentence Prediction) task and MLM pre-training has been extended to the multilingual setting by modifying MLM training to include concatenated translation pairs, known as translation language modeling (TLM).

For classification tasks, given an input sentence, its input representation is the sum of the corresponding token, segment and position embeddings. Besides, generally, we use a special token [CLS] as its sentence vector representation.

Therefore, our proposed technique utilizes the LaBSE model as the base model to encode the language features and use the first input token [CLS] to obtain the sentence vector representation. Hence, for the $i$-th sentence, the sentence vector is expressed as follows.

$$S_i = LaBSE(a_i, b_i, c_i)$$

Where $a_i, b_i, c_i$ are the token embeddings, the segmentation embeddings and the position embeddings respectively.

---

[1] https://github.com/bojone/labse

## 3.2 Language-specific words extraction

In multi-task learning, the imbalance of the amount of data among different languages leads to the general representation layer focusing more on languages with more training data. To obtain general representation, we adopt adversarial training with MLM to prevent the general representation module with a base model called LaBSE from paying too much language-specific information.

As we all know, for samples of different languages, there are certain tokens that contain a large amount of language information. To avoid the general representation layer focusing too much on specific language information, the first step is to identify words that contain rich language information, so we proposed to build a language recognition model based on the LaBSE model. Guided by supervised fitting task representation, the model can learn and distinguish language information. When the model predicts a sample, the output prediction probability can be regarded as the proportion of language information contained in the sample estimated by the model. We further use this model and MLM strategy to quantify the amount of language information of each word.

Specially, let $S = [x_1, x_2, x_3, \ldots, x_n]$ denote the input sentence, and $O_y(S)$ refers to the output prediction probability by the language recognition model LaBSE for correct label $y$. The language information $I_{w_i}$ of word $w_i$ id defined as

$$I_{w_i} = O_y(S) - O_y(S_{\backslash w_i}),$$

where $S\backslash_{w_i} = [w_0, w_1, \ldots, [MASK], \ldots, w_n]$ is the sentence after replacing $w_i$ as $[MASK]$.

Later, for each sentence, we rank all the words according to the ranking score $I_{w_i}$ in descending order and only take the score greater than 0 as the language-specific words to form a word list $I$.

## 3.3 Language information perturbation

Adversarial training presented by Goodfellow et al. (2015) is a novel regularization method that improves the robustness of misclassifying small perturbed inputs (Sato et al., 2018). Following great success in the image processing field, Miyato et al. (2017) first proposed to apply this idea to natural language processing (NLP) tasks. He pointed out that adding perturbation to the input word embedding space improves the generalization performance of models for NLP tasks. Inspired by his research, in order to make the model have greater language generalization, we further increase the adversarial perturbations for the words embedding of those language-specific words selected in section 3.2.

Specially, let $r_{AdvT}^t$ be adversarial perturbation vector for $t$-th word $x^t$ in word embedding vectors $[x^1, x^2, \ldots, x^t]$ as $x$ and $y$ represent the label for each language. We assume that $r_{AdvT}^t$ is a D-dimensional vector whose dimension always matches that of word embedding vector $w^t$. $L(x, y, \theta)$ is the loss function of individual training sample $(x, y)$ in training dataset where $\theta$ are the model parameters, then the adversarial perturbation $r_{AdvT}^t$ is calculated as follow.

$$r_{AdvT}^t = \alpha \frac{\epsilon g^t}{\|g\|_2}$$

$$g^t = \nabla_{w^t} L(x, y, \theta)$$

$$L(x, y, \theta) = \log p(y|x; \theta)$$

Where $g$ is a concatenated vector of $g^t$ for all $t$ and $\alpha$ is a weight threshold which represents the degree of perturbation in language-specific words. In this paper, the weight value $\alpha$ of language-specific word is set to 1.5 and others are set to 1.0. The optimal value of this value will be proved in Section 5.3.

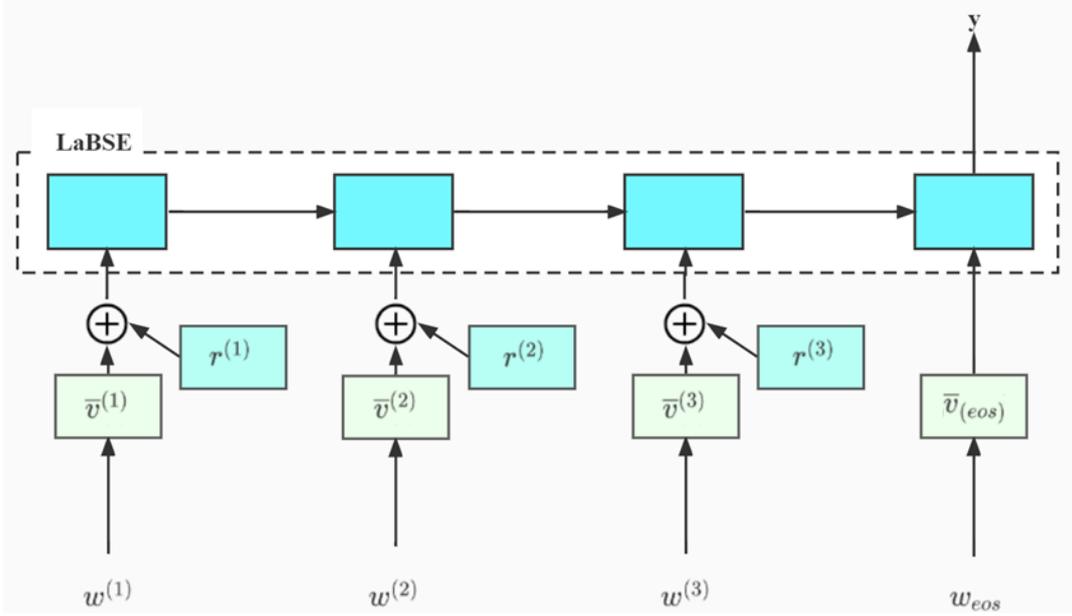

Figure 2. Adversarial training.

## 3.4 Language-specific representation

There is a certain correlation among different languages of the Dravidian language family. Therefore, in order to better recognize the interaction among different languages, we introduced language descriptors based on the self-attention mechanism proposed Vaswani et al. (2017) to simulate the interaction.

In the adversarial training process, we use the disturbing gradient for backpropagation and parameter update, then remove the noise of the embedding layer, restore the original gradient, and perform the next epoch of iterative training.

Formally, assuming that a language descriptor means one kind of language label and is represented as a vector $N_i \in \mathbb{R}^m$ where $m$ is equal to the general representation dimensionality. Consequently, all language descriptors for all languages can compose a matrix $N \in R^{n \times m}$, and $n$ is the number of languages, and each row is the descriptor for a certain language. Therefore, a language descriptor for a certain language $i$ is obtained as follows.

$$N_i^{new} = softmax(N_i N^T) N$$

We first calculate the dot product between the original descriptor $N_i$ and the other descriptors, which is then normalized by the $softmax$ function. The output vector can represent the interactions among different languages. Next, the dot product of this output and $N$ is calculated to obtain the new descriptor $N_i^{new}$, which can be considered as the weighted sum of all language descriptors with regarding to the language $i$.

## 3.5 Multi-dimension information fusion

In order to fully learn the general semantic information and the correlation among languages, we further spliced the language descriptor with the sentence vector $S_i$ output by the general representation module to fuse multi-dimensional semantic information and map it to the labels dimensions corresponding to each language by a fully connected layer.

$$h_i = [S_i; N_i^{new}]$$
$$P = (Wh_i + b)$$

Here $W$ and $b$ are parameters of the fully connected layer, $h_i$ is the spliced vector of $S_i$ and $N_i^{new}$.

## 4. Experiment

### 4.1 Dataset

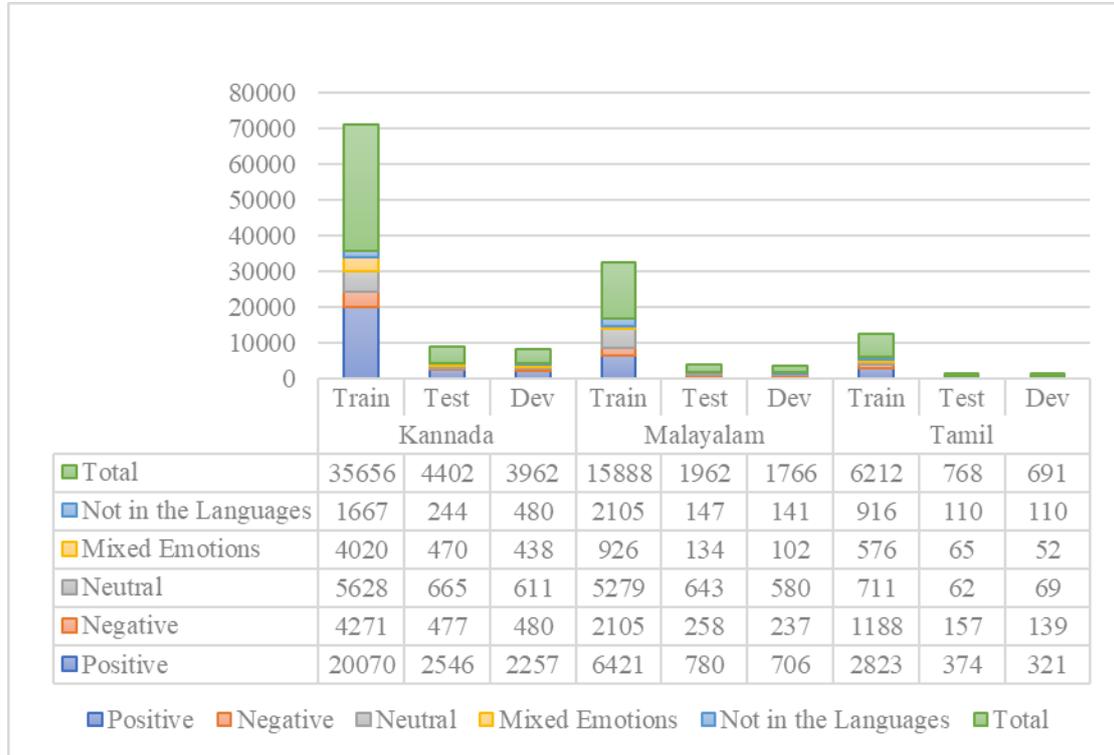

Figure 3. Data distribution for task 1.

In order to ensure the effectiveness of the framework, it has been tested on two different tasks. Task 1: Fire2021 message-level polarity classification task[1] (Priyadharshini et al., 2021). This task gives a code-mixed dataset of comments/posts in Tamil-English, Malayalam-English, and Kannada-English. Based on this dataset, the participators have to classify it into one of the five labels (positive, negative, neutral, mixed emotions, or not in the intended languages). The data distribution is shown in Figure 3. Task 2: EACL2021 Offensive language identification[2] (Chakravarthi et al., 2021). This task is to identify offensive language content of the code-mixed dataset of comments/posts in Dravidian Languages, which is shown in Figure 4.

---
[1] https://competitions.codalab.org/competitions/30642
[2] https://competitions.codalab.org/competitions/27654

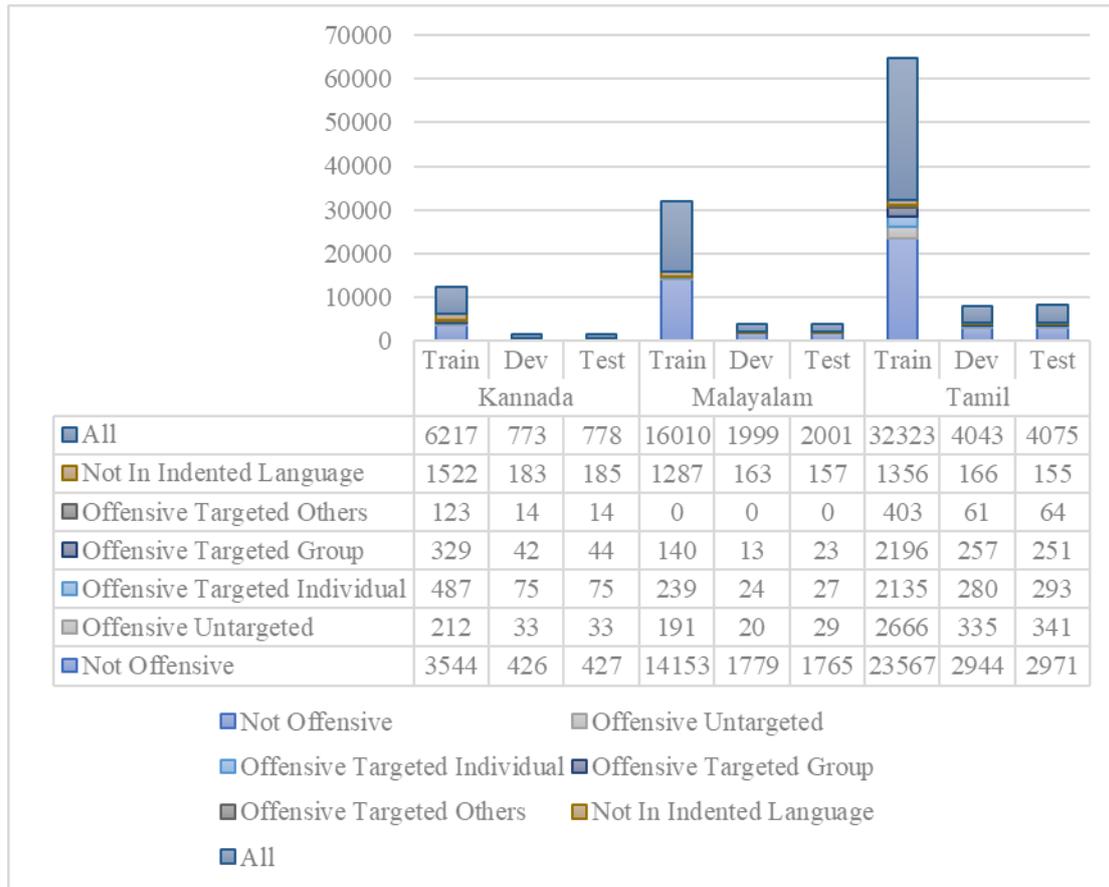

Figure 4. Data distribution for task 2.

## 4.2 Metric

Different from macro-F1, the weighted-F1 calculates metrics for each label, and finds their average weighted by the number of true instances for each label, allowing it to take into account the importance of different categories and achieving better results in the evaluation for imbalance data. Both evaluations use weighted-F1 as the evaluation index. In order to accurately compare with the evaluation teams' models, we selected weighted-F1 index as the evaluation metric.

## 4.3 Detail of experiments

In this paper, we compare our framework with the commonly used deep pre-trained model (XLM, XLM-RoBERTa, Muril, LaBSE and BERT). The relevant parameters are shown in Table 1. Besides, we use pytorch and transformers framework to implement the models.

Table 1. The parameter of the pre-trained model

| Parameter | values |
| --- | --- |
| Learning rate | 5e-5 |
| Dropout | 0.5 |
| Weight decay | 0.001 |
| Optimization function | Adam |
| The largest number of the sentence | 128 |
| Batch size | 64 |

## 5. Results and analysis

## 5.1 Comparative experiments

In order to verify the effectiveness of our framework, we designed 7 sets of experiments in this paper, consists of 3 sets of model comparison experiments, 3 sets of ablation experiments and 1 set of exploration experiment. As for the comparison experiments, to select a high-performance classification model, we explore the performance among different deep learning models, the performance among pre-trained models and the performance between multilingual training and monolingual training. Besides, we also compare the performance of our framework with the state-of-the-art model. In light of ablation experiments, it is mainly to verify the effectiveness of each module in our framework. Moreover, we further explore the weight setting of α during language information perturbation module.

**Comparative experiment 1: Performance comparison among different deep learning models.** In order to select the base model, we compare the performance during pre-trained models such as XLM-RoBERTa-Base (Conneau et al., 2020), XLM (Conneau et al., 2020), LaBSE, Muril (Khanuja et al., 2021) and Multilingual-BERT (Devlin et al., 2019) on the FIRE 2021. The results are shown in Figure 5, it is obvious to see that the LaBSE model is relatively effective in each task in terms of average scores.

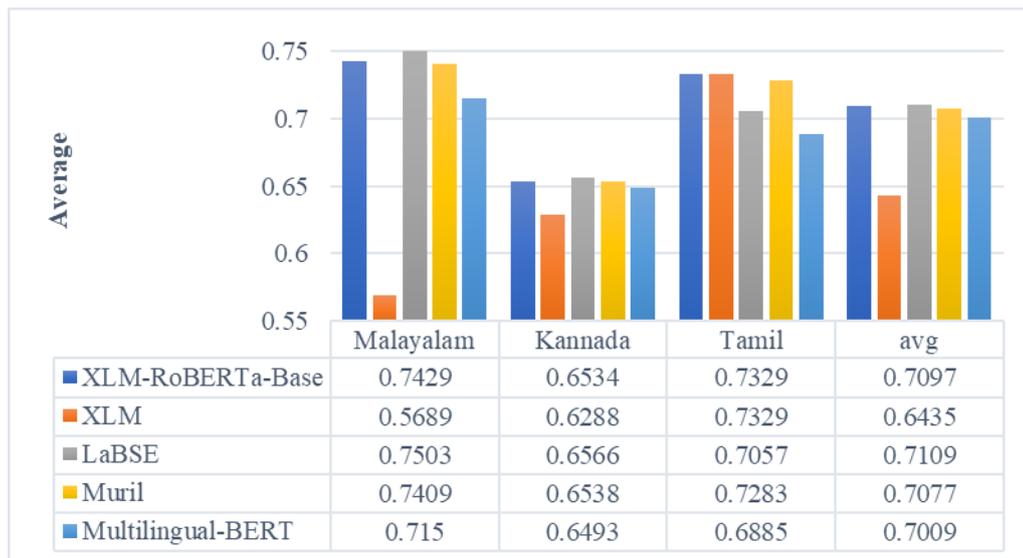

Figure 5. The results of pre-trained models.

**Comparative experiment 2: Performance comparison between the model trained with multilingual corpus and one trained with monolingual corpus.** After selecting the best base model, we further compared the performance of the model trained with multilingual corpus and one trained with

monolingual corpus on the FIRE 2021 dataset and EACL 2021 dataset. As shown in Table 2, the model trained with monolingual corpus outperforms that one trained with multilingual corpus. We explored the reason for this result is that the general representation module pays too much attention to the language-specific information, leading to a decline in the generalization capability of the model.

Table 2. The results of multilingual corpus training.

| Task | Language | Monolingual | Multilingual |
|---|---|---|---|
| FIRE 2021 | Malayalam | 0.7503 | 0.7388 |
| | Kannada | 0.6566 | 0.6494 |
| | Tamil | 0.6957 | 0.7065 |
| EACL 2021 | Malayalam | 0.9677 | 0.9606 |
| | Kannada | 0.7854 | 0.7813 |
| | Tamil | 0.8120 | 0.8201 |
| Average | | **0.7780** | 0.7761 |

**Comparative experiment 3: Performance comparison with the state-of-the-art methods.** In this paper, the datasets we used are from FIRE 2021 and EACL 2021 evaluation competitions. Therefore, in order to verify the effectiveness of the proposed framework, we further compare the results of our framework with the top 10 teams in these two evaluation competitions (shown in Figure 6 and Figure 7). Since the task in our paper is a multilingual task, in Figure 5 and Figure 6 the ranking is based on the average of each team in the three languages. The experimental results show that our framework achieved the best results in most languages of the task (see the red line).

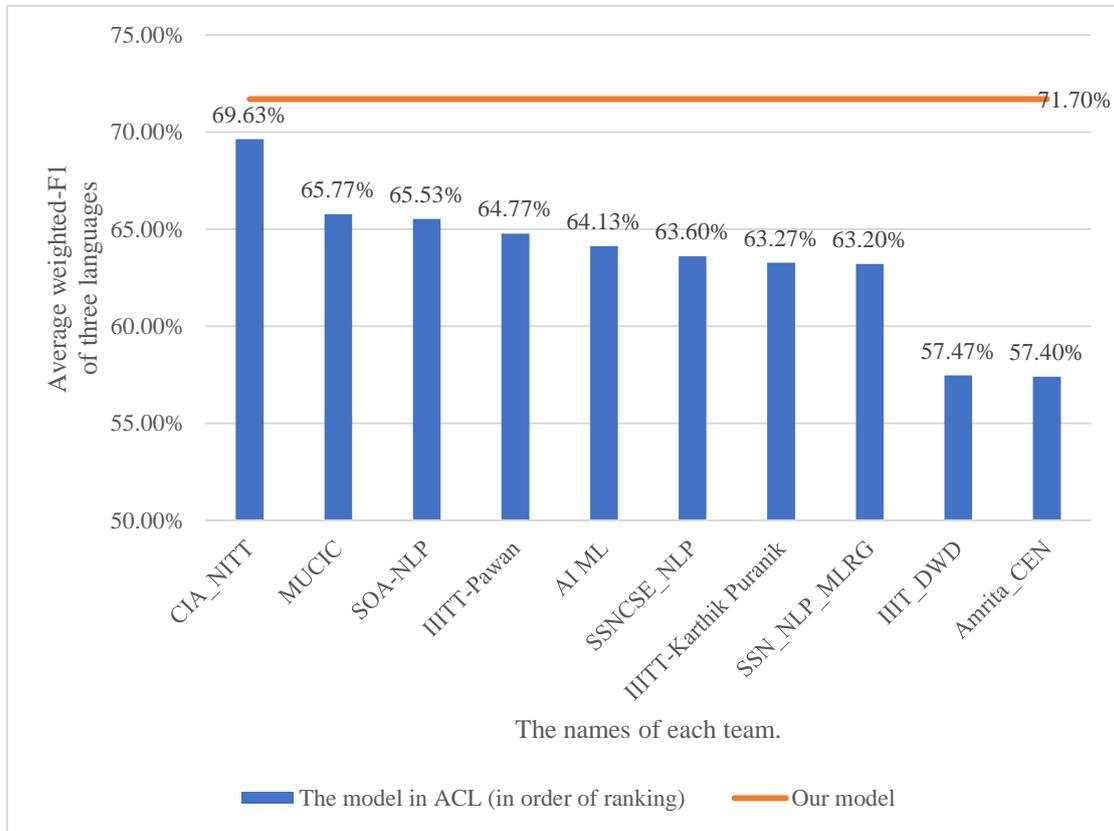

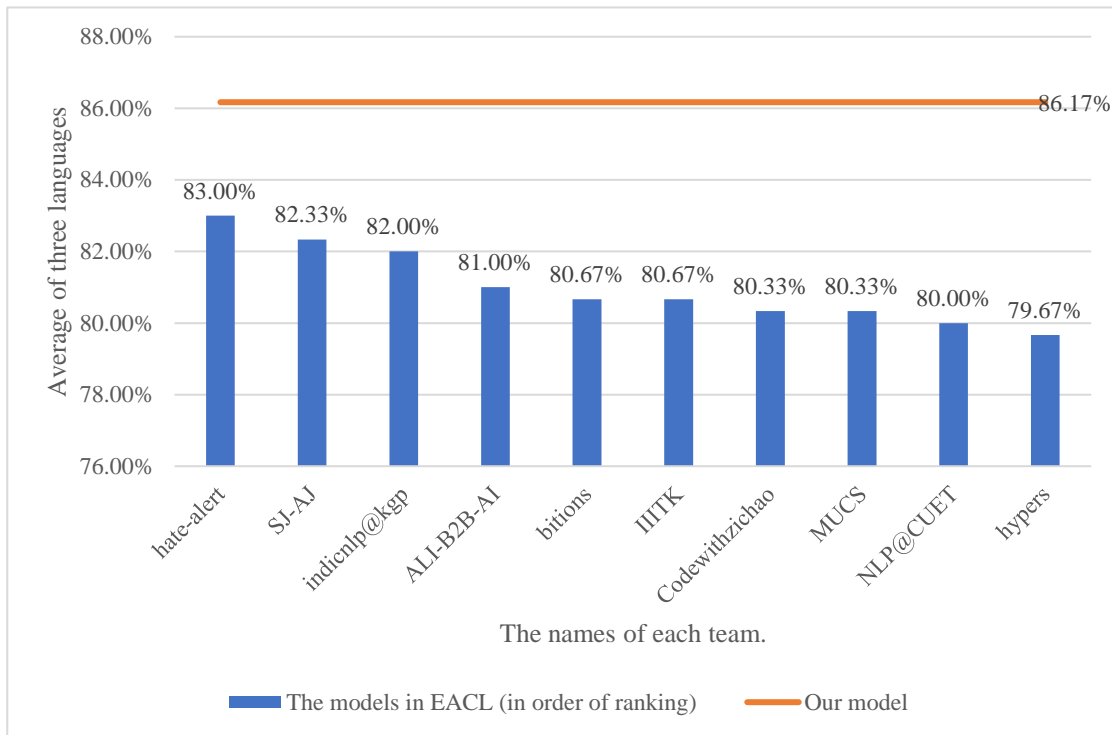

Figure 6. The results on FIRE 2021 dataset.

Figure 7. The results on EACL 2021 dataset.

We also compared our model with the state-of-the-art language model of each language on the two datasets. The results are shown in Table 3. In the Malayalam task and Tamil task of FIRE 2021 and the Malayalam task of EACL 2021, our model did not exceed the monolingual language state-of-the-art model. But from the average result, our model is better than the combination of three state-of-the-art monolingual languages, which is an increase of 0.2% on FIRE 2021 and an increase of 2.84% on EACL 2021.

Table 3. Comparison of our model with the state-of-the-art monolingual language model.

| Dataset | Model | Malayalam | Kannada | Tamil | Avgeage |
|---|---|---|---|---|---|
| FIRE 2021 | State-of-the-art | **0.8040** (ZYBank-AI Team) | 0.6300 (SSNCSE_NLP) | **0.7110** (CIA_NITT) | 0.7150 |
| | Ours | 0.7599 | **0.6822** | 0.7083 | **0.7170** |
| EACL 2021 | State-of-the-art | **0.9700** (hate-alert) (Saha et al., 2021) | 0.7500 (SJ-AJ) (Jayanthi and Gupta, 2021) | 0.7800 (hate-alert) (Saha et al., 2021) | 0.8333 |
| | Ours | 0.9637 | **0.7910** | **0.8314** | **0.8617** |

## 5.2 Ablation experiment

**Ablation experiment 1: Effectiveness verification of adversarial training.** In this paper, on the basis of selecting the LaBSE model as the base model to obtain sentence vectors, adversarial training was added to the word embedding layer for joint training. We utilize the FGM algorithm to regularize the word embedding layer of the LaBSE model and perturb it to enhance the robustness of the model. On the FIRE 2021 dataset, we tried to add adversarial training to the monolingual models and the multilingual model. As the result shown in Table 4. It is obvious to see that adding perturbation to the word embedding layer of the model can improve the performance to a certain extent. Because adding perturbation to the model is equivalent to generating adversarial samples to the model, which enhances the diversity of training samples. At the same time, the overall performance of the multilingual model that only uses adversarial learning is not as good as training three monolingual models separately.

Table 4. Comparison of adversarial training.

| Model | Malayalam | Kannada | Tamil | Average |
| --- | --- | --- | --- | --- |
| Monolingual | 0.7503 | 0.6566 | 0.6957 | 0.7009 |
| Monolingual + adversarial training | 0.7543 | 0.6771 | 0.7007 | **0.7107** |
| Multilingual | 0.7388 | 0.6494 | 0.7065 | 0.6982 |
| Multilingual + adversarial training | 0.7473 | 0.6545 | 0.7018 | 0.7012 |

**Ablation experiment 2: Effectiveness verification of language-specific representation module.** In order to extract language-specific words, on the basis of adversarial training, we further proposed to train a language recognition model and apply MLM strategy to quantify the amount of language information of each word. We verify the performance of this module on the FIRE 2021 dataset. As shown in Table 5, the experimental results demonstrate that this module has a certain impact on classification performance with the average value improving 1.13% in terms of comparing with the model without this module, which indicates that this module can make the general representation module more unbiased to increases the generalization ability of the model.

Table 5. Comparison of language-specific representation module.

| Model | Malayalam | Kannada | Tamil | Average |
| --- | --- | --- | --- | --- |
| Multilingual | 0.7388 | 0.6494 | 0.7065 | 0.6982 |
| Multilingual + adversarial training | 0.7473 | 0.6545 | 0.7018 | 0.7012 |
| Multilingual + adversarial training + language-specific words extraction | 0.7567 | 0.6849 | 0.6958 | **0.7125** |

**Ablation experiment 3: Effectiveness verification of language descriptor.** Since all the corpus we used belongs to the Dravidian language family, there is a certain correlation among them. Therefore, in order to integrate language-related information into the model, we proposed to introduce language descriptors for joint training during the language representation module. The results are shown in Table 6. In terms of average score, the classification result increased by 0.45% compared to the experimental results without this module, verifying the effectiveness and feasibility of this framework.

Table 6. Comparison of language descriptor.

| Model | Malayalam | Kannada | Tamil | Average |
| --- | --- | --- | --- | --- |
| Multilingual | 0.7388 | 0.6494 | 0.7065 | 0.6982 |
| Multilingual + adversarial training | 0.7473 | 0.6545 | 0.7018 | 0.7012 |
| Multilingual + adversarial training + language-specific words extraction | 0.7567 | 0.6849 | 0.6958 | 0.7125 |
| Multilingual + adversarial training + language-specific words extraction + language descriptor | 0.7599 | 0.6822 | 0.7083 | **0.7170** |

## 5.3 Explore experiments

Eventually, the value of weight α is also verified in this paper. We explored the influence of this parameter on the multilingual model based on adversarial training. The results shown in Table 7 demonstrate that when the value of α is 1.5, the model performs the best. This means that the perturbation degree of language-specific words by the model is 1.5 times than the perturbation degree of other words.

Table 7. The results of threshold.

| Threshold | Mal | Kannada | Tamil | Average |
| --- | --- | --- | --- | --- |
| 1.1 | 0.7562 | 0.6655 | 0.7099 | 0.7105 |
| 1.2 | 0.7528 | 0.6668 | 0.6917 | 0.7038 |
| 1.3 | 0.7526 | 0.6685 | 0.6836 | 0.7016 |
| 1.4 | **0.7605** | 0.6635 | 0.6953 | 0.7064 |
| 1.5 | 0.7567 | **0.6849** | **0.6958** | **0.7125** |

## 6. Conclusion

In this paper, we proposed a multilingual text classification framework for the Dravidian languages. On the one hand, the framework uses the LaBSE pre-trained model as the base model. Aiming at the problem of text information bias in multi-task learning, we proposed to use the MLM strategy to select language-specific words, and implemented adversarial training to perturb them. On the other hand, in view of the problem that the model cannot well recognize and utilize the correlation among languages, we further proposed a language representation module to enrich semantic information for the model. In the future, we will expand the number of languages and improve the performance of the multilingual text classification framework.